\documentclass[a4paper,twoside]{article}

\usepackage{epsfig}
\usepackage{subcaption}
\usepackage{calc}
\usepackage{amssymb}
\usepackage{amstext}
\usepackage{amsmath}
\usepackage{amsthm}
\usepackage{multicol}
\usepackage{pslatex}
\usepackage{apalike}
\usepackage{algorithm2e}
\usepackage[bottom]{footmisc}
\usepackage{SCITEPRESS}     

\begin{document}

\title{Game State and Spatio-temporal Action Detection in Soccer using Graph Neural Networks and 3D Convolutional Networks}

\author{\authorname{Jérémie Ochin\sup{1}\sup{2}, Guillaume Devineau\sup{2}, Bogdan Stanciulescu\sup{1} and Sotiris Manitsaris\sup{1}}
\affiliation{\sup{1}Centre for Robotics, MINES Paris - PSL, France}
\affiliation{\sup{2}Footovision, France}
\email{jeremie.ochin@minesparis.psl.eu, bogdan.stanciulescu@minesparis.psl.eu, sotiris.manitsaris@minesparis.psl.eu, guillaume.devineau@footovision.com}
}


\abstract{Soccer analytics rely on two data sources: the player positions on the pitch and the sequences of events they perform. With around 2000 ball events per game, their precise and exhaustive annotation based on a monocular video stream remains a tedious and costly manual task. While state-of-the-art spatio-temporal action detection methods show promise for automating this task, they lack contextual understanding of the game. Assuming professional players' behaviors are interdependent, we hypothesize that incorporating surrounding players’ information such as positions, velocity and team membership can enhance purely visual predictions. We propose a spatio-temporal action detection approach that combines visual and game state information via Graph Neural Networks trained end-to-end with state-of-the-art 3D CNNs, demonstrating improved metrics through game state integration.}

\onecolumn \maketitle \normalsize \setcounter{footnote}{0} \vfill

\section{\uppercase{Introduction}}
\label{sec:introduction}

Sports Analytics, the practice of collecting and interpreting data from past performances to inform future decisions, was pioneered in soccer by Royal Air Force Wing Commander Charles Reep, who meticulously analyzed over 600,000 passing moves beginning in the 1950s \cite{Mandadapu2024}. Since then, the demand for such statistics has experienced a strong development with the increased ability to collect, process, store, and analyze large quantities of data, using, among other things, cameras, GPS, computers, and algorithms \cite{Jha2022}. However, if recent advances in deep learning have facilitated the automation of player tracking and pitch location from monocular videos - a convenient alternative to the wearable -, the task of precise event annotation from soccer videos is still performed manually by trained operators \cite{Cartas2022}.

\bigskip

\begin{figure}[h!]\centering
\includegraphics[width=.99\linewidth]{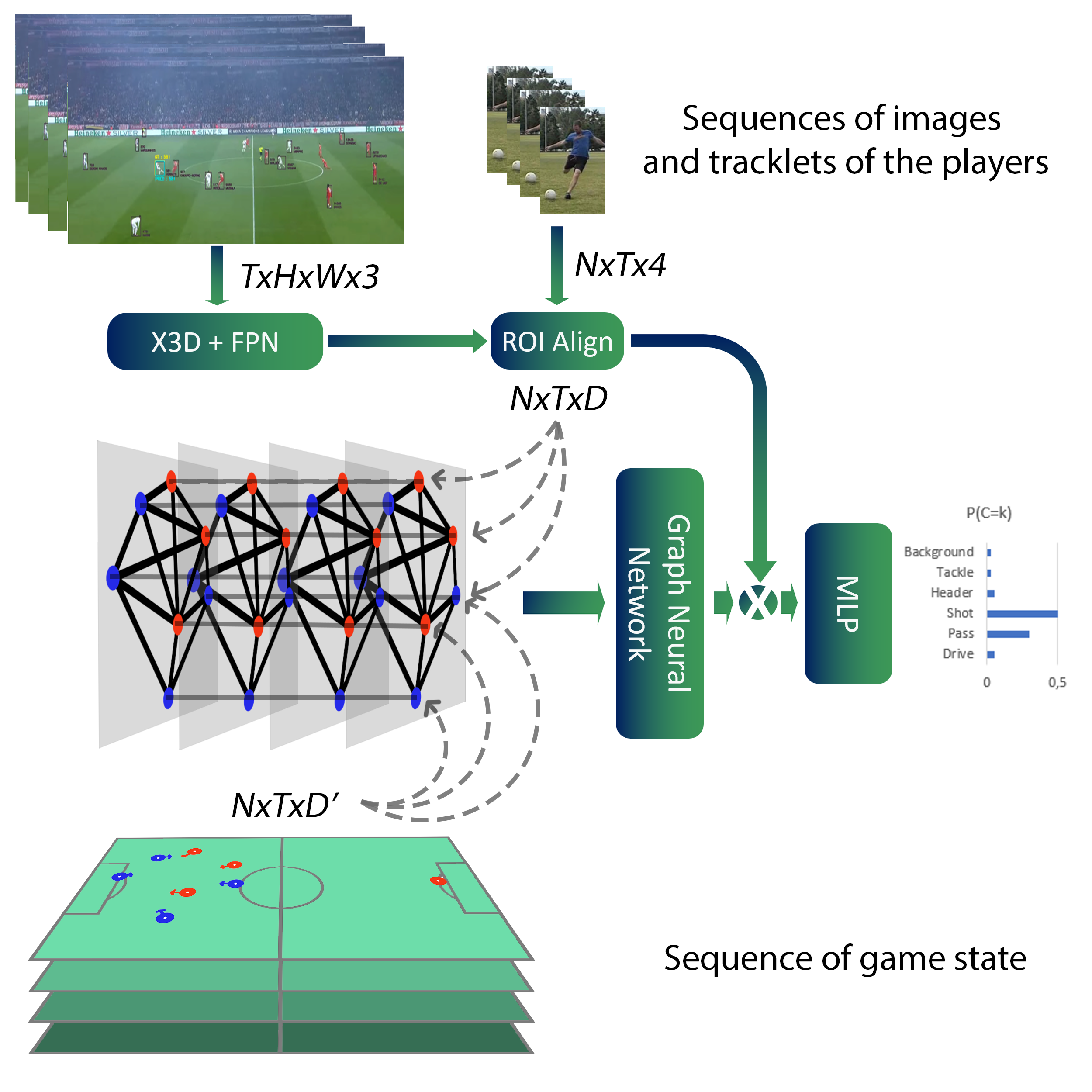}
\caption{Spatio-temporal Action Detection in soccer, using a Graph Neural Network to encode Local Game State and a Track-Aware Action Detector \cite{Singh2022} to extract relevant visual features. Our method demonstrates the complementarity of game state information and visual features and leads to improved performances.}
\label{fig:schema_global}
\end{figure}

Indeed, sport video understanding remains a highly challenging domain that encompasses a wide variety of tasks and applications. This article specifically addresses the spatio-temporal localization of events in soccer games. The goal is to accurately identify, in untrimmed videos, \textbf{when} an action occurs, \textbf{what} the action is, and \textbf{who} performed it, or \textbf{where} it took place since an action instance is defined as a set of linked bounding boxes over time, called action tube or tubelet, that contains the player performing the action. Solving this problem is essential for generating reliable and actionable statistics that can inform tactical decisions and allow opponent analysis or performance evaluations in professional soccer.

\bigskip

Current methods of spatio-temporal action detection are not designed to explicitly capture the complex relationships that can be observed in soccer gameplay. Soccer is a highly interactive sport, where player actions are decided based on the positions, movements of both teammates, opponents and their collective strategy. Existing methods, which rely on the aggregation of local visual features along the path of the players in screen space, lack the ability to fully comprehend these contextual relationships.

\bigskip

Jointly inspired by prior works in spatio-temporal action detection, group activity recognition and graph-based methods for soccer action spotting, our work proposes to enhance a track-aware spatio-temporal event detection method (TAAD) \cite{Singh2022} by incorporating, for each player that is observed, contextual information about the local game state, through a learnt node embedding that is produced by a Graph Neural Network (GNN) (see Figure \ref{fig:schema_global}). We hypothesize that adding this information about the surrounding players, such as their positions and velocities on the pitch, team membership and appearance, structured by a GNN, can improve the performance of pure "visual" event detection models.

\bigskip

It is worth noting that using TAAD alongside game state information aligns naturally with soccer analytics, where tracking, identification, and position estimation are prerequisites for event annotation. Since tracklets and game state details are typically available before event annotation begins, it is natural to leverage them to enhance event detection.

\bigskip

We test our hypothesis by developing a new spatio-temporal action detection dataset dedicated to soccer, and by using a GNN trained end-to-end alongside a state-of-the-art 3D convolutional neural network (CNN), implementing a TAAD architecture. Given the applied nature of this research and the goal of enabling concrete applications, we aim to limit computational costs by ensuring that this implementation is trainable on a single entry-level professional GPU, such as an RTX A6000.

\bigskip

Our results demonstrate that explicitly incorporating structured game state information improves the performance of action detection models.

\section{\uppercase{Related Work}}

\subsection{Spatio-temporal Action Detection}

In the field of video understanding, spatio-temporal action detection is the task of detecting intervals of possibly multiple and concurrent human actions in untrimmed videos. State-of-the-art methods are deep learning based and most of the work in this field addresses several core challenges:

\begin{itemize}
\item identify the actors, 
\item optionally model the relationship between them or with contextual elements, 
\item generate relevant features from the frames or video, 
\item sample and aggregate these features in a relevant manner, 
\item produce the frame-by-frame or clip-level action predictions, 
\item temporally link these predictions to obtain an action tube.
\end{itemize}

These methods are divided into two main categories: frame-level methods and clip-level methods.

\bigskip

\textbf{Frame-level methods:} the predictions are made frame by frame and then linked together to form action tubes. Conceptually, most of these methods rely on a region proposal network and ROI Pooling or ROI Align layers \cite{MaskR2018} to (i) detect the actors and (ii) locally sample and pool the features that are produced by a CNN. Early methods used 2D CNNs and optical flow for motion cues \cite{Weinzaepfel_2015,Saha_2016,Singh_2017}, later transitioning to 3D CNNs for enhanced feature extraction \cite{gu_ava_2018,Girdhar_2018}. Expanding on this line of work, Track-Aware Action Detector (TAAD) first detects and tracks actors, using modern object detectors and tracking algorithms - which later facilitates the linking of predictions -, then aggregates features along the path for robustness to camera motion \cite{Singh2022}. Finally, some approaches integrate actor localization and action prediction, resembling YOLO's object detection approach \cite{Yolo2016,chen_watch_2021,Yowo2021}.

\bigskip

Per-frame detections are linked using methods like dynamic programming with cost functions based on detection scores and overlaps \cite{Saha_2016,Singh_2017,kalogeiton_action_2017} or tracking-by-detection \cite{Weinzaepfel_2015}.

\bigskip

\textbf{Clip-level methods:} these methods predict tubelets, sequences of bounding boxes that tightly bind the actions of interest. The Action Tubelet Detector (ACT-detector) uses 2D CNNs to stack per-frame features for regressing the offset and dimensions of a cuboid anchor and predict the action class \cite{kalogeiton_action_2017}. Similarly, another approach estimates action centers and build cuboid anchors around them \cite{li_actions_2020}. More recently, a Transformer-based method "encodes" video data using self-attention layers and "decodes" tubelets locations, classes, and scores using learned "tubelet queries" and cross-attention layers \cite{zhao_tuber_2022}.

\bigskip

\textbf{Visual relationship modeling:} the interaction between actors, objects, and their environment provide valuable context for improving detection and classification in videos. Several approaches have been proposed to model these relationships, such as attention mechanisms or GNNs. 

\bigskip

Attention-based, the Actor Transformer Network \cite{girdhar_video_2019} employs a region proposal network to sample features from a 3D CNN, creating actor representations. Then, the attention mechanism attend to all other spatio-temporal features extracted from the video, adding contextual information. Alternatively, attention can focus solely on actor tokens and a fixed-size environment representation obtained via average pooling of the feature map \cite{vo_agent-environment_2021}.

\bigskip

GNNs are also popular for modeling relationships, passing information between nodes representing entities and their relationships. Often, these methods first detect and track actors, sample features from 3D CNNs, and use these as node features, while edges connect actors, objects, or temporal sequences \cite{zhang_structured_2019,tomei_video_2021}. Action predictions can occur on edges using embeddings of interacting entities or at the node level with entity-specific embeddings.

\bigskip

\textbf{Datasets:} most of the works described above were benchmarked on the UCF101-24 \cite{UCF2012} or AVA \cite{gu_ava_2018} datasets. They contain material that is substantially different from sports videos, where fast camera motions can happen, actors appear small on screen, with motion blur and, in team sports, with fluctuating visibility. Therefore, to foster the development of methods adapted to these characteristics, a dedicated multi-person video dataset of spatio-temporally localized sports actions was created: MultiSports \cite{Multisports2021}. It contains 3200 video clips collected from 4 sports (basketball, volleyball, soccer and aerobics gymnastics) and provides action tubelets annotations for these videos.

\subsection{Group Activity Recognition in sports} 

Group activity recognition consists in reasoning simultaneously over multiple actors in order to predict their individual and group activities. Some of the methods draw inspiration from visual relationship modeling, presented above. We will present only a few methods that tackle individual and group activity recognition in Volleyball.

\bigskip

In the early work Social Scene Understanding \cite{bagautdinov_social_2016}, a 2D CNN backbone is shared for player detection and activity recognition. A first stream produces candidate players detections, filtered using a Markov Random Field. Then, players features are sampled to produce fixed-size vector representations per frame. These are fed into an Recurrent Neural Network (RNN), whose hidden states enable individual action predictions frame-by-frame, while max pooling these states across players yields group activity predictions. Similarly, the Convolutional Relational Machine \cite{azar_convolutional_2019} uses 2D or 3D CNN features to generate "activity maps"— heatmaps indicating the likelihood of individual and group actions per class. Both approaches rely on the wide receptive field of CNNs for multi-agent reasoning, though this may lack the structural expressiveness offered by attention mechanisms or graphs.

\bigskip

Proposed more recently, the Actor-Transformer for Group Activity Recognition \cite{gavrilyuk_actor-transformers_2020} integrates RGB, Optical Flow, and Pose modalities. Players are "tokenized" via embedding features (using ROI Align) or pose coordinates, which serve as inputs to a Transformer encoder for individual and group action predictions. The centers of the bounding boxes in screen space are used as positional encoding of the players.

\bigskip

While both Visual Relationships Modeling methods and Group Activity Recognition methods effectively model relationships between entities within an image or video, they differ from approaches that leverage game state information, such as player positions, velocities, team membership and other contextual data, that are better suited to understand the complex contextual relationships that drive gameplay.

\subsection{Graph Neural Networks in sports}

GNNs are ubiquitous. Their ability to model interactions within structured data has made them good candidates to model inter-agent patterns in a number of multi-agents problems. This section highlights methods applied to sports, particularly soccer, that use tabular player data (the "game state") as node representations rather than visual features, aiming to learn new discriminative features that could complement the visual ones.

\bigskip

In \cite{Everett2023}, a combination of GNN and LSTM is used to impute football players' locations at all time steps from sparse event data (e.g., shots, passes). Similarly, \cite{Ding2020} employs a graph-attention network and dilated causal convolutions to predict trajectories of coordinated agents, such as basketball and football players. These studies highlight the strong link between players' spatial configurations and actions, as well as the ability of GNNs to deliver robust structured predictions.

\bigskip

Lastly, in \cite{Cartas2022}, a method for temporal event detection in football leverages player positions, velocities, and team identities (team 1, team 2, goalkeepers, referees). A GNN with edge convolutions and graph pooling encodes player configurations into single vectors for each time step, which are then used for action detection. Building on this approach, we propose a spatio-temporal action detection method that removes the graph pooling step to enable node-based predictions (per player predictions).

\subsection{Vision-Language Models} 

Vision-Language Models (VLMs) integrate large language models with computer vision capabilities, enabling them to understand and generate text based on visual inputs. Recent notable examples include Molmo, Pixtral, LLaVa and Qwen2-VL \cite{2409.17146,2410.07073,2304.08485,2409.12191}. However, VLMs are not yet mature enough for soccer video analysis because they rely heavily on proprietary models and synthetic data, lack specialized training datasets tailored to soccer, struggle to interpret complex player interactions and tactics, and tend to hallucinate - a critical flaw in the sensitive domain of sports analysis where accuracy is paramount.

\section{\uppercase{Methodology}}

\subsection{Dataset}

\begin{figure*}[h]\centering
\includegraphics[width=.99\linewidth]{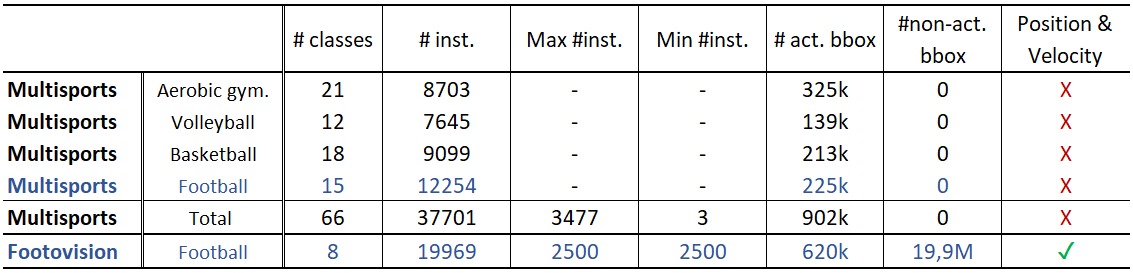}
\caption{Comparison of the characteristics of the MultiSports and Footovision datasets}
\label{fig:compdatasets}
\end{figure*}

The MultiSports dataset is a reference benchmark for multi-person spatio-temporal action detection methods in sports, but it is missing important features for this work: it lacks detection and tracking data of all the players as well as camera calibration models in order to estimate the positions of the players on the field. These are required features to study if game state information can help enhancing pure visual spatio-temporal event detection methods. In order to remedy these issues, we created a specific "ball events" dataset that matches these requirements (see Figure \ref{fig:compdatasets}). The \textbf{Footovision dataset} characteristics are:

\bigskip

\begin{itemize}
    \item\textbf{20000 videos} of at least 75 frames (3 seconds). 
    \item\textbf{997 different football games} from all over the world.
    \item\textbf{2500 samples per class} minimum. 
    \item\textbf{8 classes of action:} ball-drive, pass, cross, header, throw-in, shot, tackle and ball-block
\end{itemize}

\bigskip

Sequences are sampled around randomly selected events from 997 games, ensuring no overlap between clips to create distinct training and validation sets. Some classes exceed 2,500 instances due to the high likelihood of certain events preceding or following others in a short period of time (e.g., ball-drives often lead to passes, crosses, or shots and are therefore present when a pass, cross or shot is sampled). This introduces a slight class imbalance.

\bigskip

The data used to make the annotations was collected from TV broadcast videos by the company Footovision, using in-house annotation software and team of professional annotators. Multiple manual quality controls are performed on tracking and identification data, line detection and camera calibration, and events are manually annotated with an estimated average precision of a few frames around the "ball touch". Ball-drives are annotated from the first ball touch to the next action (pass, cross, shot, tackle). All other events are annotated in a window of 7 frames around the ball touch. These annotations contain, frame by frame, for each video clip:
\bigskip
\begin{itemize}
    \item the bounding box of the players and referees, with a tracklet ID to match bounding boxes across the clip (for the avoidance of doubt : this dataset does not contain the ball position),
    \item the team membership of the players (team 1 or 2),
    \item their shirt number,
    \item the class of the action performed by the player, including a background class when the players do not perform any action,
    \item their position and velocity in the pitch reference frame.   
\end{itemize}
\bigskip
The videos in this dataset are subject to copyright protection and cannot be shared publicly.

\subsection{Visual feature extraction}

\subsubsection{Selection of a baseline method}

We selected the Track-Aware Action Detector (TAAD) \cite{Singh2022} as our baseline due to its suitability for sports with significant camera motion. TAAD tracks players and aggregates features from a state-of-the-art 3D CNN along the track using ROI Align \cite{MaskR2018}, followed by a Temporal Convolutional Network (TCN) for per-frame action predictions. It performs well on the MultiSports dataset and aligns with this work, as player tracking is already completed before event annotation. We aim to compare TAAD with a hybrid approach combining visual features and local game state features learned by a GNN.

\subsubsection{Players visual feature extraction}

The original TAAD implementation uses a SlowFast backbone \cite{feichtenhofer_slowfast_2019}, a state-of-the-art CNN for video understanding. Nevertheless, recent works \cite{martin2020fine,hong2022spotting,cioppa2024soccernet} demonstrate that lightweight architectures are particularly well-suited for small datasets of sports action recognition and tend to delay overfitting. For our implementation, we selected the X3D backbone family, which delivers performance comparable to SlowFast while using a fraction of the parameters. For instance, X3D-M achieves 76\% top-1 accuracy with just 3.8M parameters, compared to SlowFast 4×16 R50, which achieves 75.6\% top-1 accuracy with 34.4M parameters on the Kinetics-400 dataset \cite{feichtenhofer_x3d_2020}. This substitution reduces computational costs for subsequent end-to-end training as we plan to extend these architectures by using a GNN.

\bigskip

We utilize the first five blocks of the X3D backbone and integrate a Feature Pyramid Network (FPN) with the last three blocks, following \cite{lin_feature_2017,Singh2022}. Player features are extracted using ROI Align based on their tracklet data, maintaining the feature dimension (D=192) from X3D's fifth block. Additionally, this backbone preserves the temporal resolution, ensuring that a clip of T frames produces a feature tensor with a temporal dimension of T.

\bigskip

After applying TAAD to our clip of T frames, with N players, we obtain a feature tensor $\Phi_{X3D} \in \mathbb{R}^{NxTx192}$, and we note $\Phi^{(i,t)}_{X3D}$ the visual feature vector of the player i and time t:

\begin{equation}\label{eq1}
\Phi^{(i,t)}_{X3D} = \textbf{ROI}(\textbf{X3D}(sequence), bbox^{(i,t)})
\end{equation}

\bigskip

\textit{Where ROI is the ROI Align sampling operation, X3D the application of the backbone on the sequence of images and  $bbox^{(i,t)}$ is the bounding box of player i at time t.}

\subsubsection{Trade-off between image size, length and frame-rate}

Down-sampling the video resolution to 640x352 pixels provides a good balance for TAAD, as further reduction degrade performance, especially on wide fields of view where players appear small. No temporal sub-sampling was applied, as most player actions in the dataset are fine-grained and occur almost "instantly" \cite{hong2022spotting}. The 25 FPS rate aligns with the average event frequency (1 event every 65 frames) and accommodates the variability in event intervals: a ball drive can last for more than 10 seconds while a sequence of events such as shot and ball block can happen in a window of 10 frames or less. The clip length of 50 frames was chosen based on these resolutions and the available computational budget.

\subsection{Local game state encoding}

\subsubsection{Game state information}

Our approach hypothesizes that incorporating structured information about players' configuration on the pitch, referred to as the game state, complements purely visual methods like TAAD. In professional soccer, players' actions, locations and velocities are closely linked to the behavior of others, team membership, and collective strategies. Since inter-class variations in fine-grained action recognition are often subtle, modeling these player relationships can aid in learning additional discriminative features.

\bigskip

Moreover, we argue that using 2D positions and velocities on the pitch is a natural choice to model these inter-agents relations in a sport such as soccer. In contrast, methods for visual relationship modeling and group activity recognition rely on attention mechanisms that focus on visual features, lacking information about the spatial arrangement of the actors, or incorporating it only through screen space coordinates when using positional encoding. All the works that aim to model the game itself - estimate the probability of passes, next actions, detection of tactical patterns, trajectory forecasting... - are based on this very synthetic representation of the game state \cite{brefeld_probabilistic_2019,fadel_contextual_2021,martens_space_2021,rudolph_modeling_2022,dick_who_2022,anzer_detection_nodate}. In this paper, we adopt a hybrid approach, using both visual features and game state information:

\bigskip

We will call $\textbf{x}^{i}_{t}$ the feature vector of player i at time t, obtained by concatenating:

\bigskip

\begin{itemize}
    \item its normalized position $(p^{(i,t)}_{x}, p^{(i,t)}_{y})$ at time t,
    \item its normalized velocity $(v^{(i,t)}_{x}, v^{(i,t)}_{y})$ at time t,
    \item its team membership, either 0 or 1,
    \item a projection of its visual feature vector $\Phi^{(i,t)}_{X3D}$ in a space of lower dimension (D'=64), noted $\varphi^{(i,t)}_{X3D}$.
\end{itemize}

\begin{equation}\label{eq2}
\textbf{x}^{i}_{t} = \left(p^{(i,t)}_{x}, p^{(i,t)}_{y}, v^{(i,t)}_{x}, v^{(i,t)}_{y},0|1,\varphi^{(i,t)}_{X3D}\right)
\end{equation}

\subsubsection{Graph of the game}

To be able to learn a node embedding meaningful for our problem, we build a spatio-temporal graph of the game $\mathcal{G}=(\mathcal{V}, \mathcal{E})$ following these steps :

\bigskip

\begin{itemize}
    \item We create a node for each player at each time-step $\mathcal{V} = \bigl\{\textbf{x}^{i}_{t}, i \in [1,22], t \in [1,T]\bigl\}$
    \item We create temporal edges between nodes of the same player at neighboring time-steps
    \item We create edges between nodes of different players at a given time-step using simple rules, such as the M closest players or using distance-based thresholds. In this implementation, we used the 6 closest players.
\end{itemize}

\subsubsection{Edge Convolutions}

A GNN can be described as a process by which the information contained in a graph evolves using functions approximated by neural networks. More precisely, it is the process by which the information contained in a node of that graph is modified by the information contained in the nodes of its neighborhood: vector messages are exchanged between connected nodes and updated using neural networks. This defining feature of GNN is called \textit{neural message passing}.

\bigskip

In a GNN, layers of message passing are performed, and at each iteration \textit{k}, the \textit{hidden embedding} $h^{(k)}_{u}$ of a node \textit{u} $\in \mathcal{V}$ is updated according to information \textit{aggregated} from \textit{u}'s graph neighborhood $\mathcal{N}(\textit{u})$.

\bigskip

Among the large variety of message passing methods, we chose Edge Convolutions with an asymmetric edge function to implement our GNN \cite{wang_dynamic_2019}. Its update method can be summarized by equation \ref{eq3}:

\begin{equation} \label{eq3}
h^{(k+1)}_{u} = MAX\left(MLP\left(h^{(k)}_{u} \middle\| h^{(k)}_{v} - h^{(k)}_{u}\right)\right)
\end{equation}

\textit{Where MLP is a shallow Multi-layer Perceptron and MAX a channel-wise max pooling operation.}

\bigskip

This choice is driven by our will to build a node embedding that explicitly combines global information about the node, captured by $h^{(k)}_{u}$, with relative and local neighborhood information, captured by $h^{(k)}_{v} - h^{(k)}_{u}$. This is the reason why we call this node embedding the "local game state encoding".

\bigskip

Its benefit is particularly salient on the first layer of convolution (k=0) where $h^{0}_{u} = \textbf{x}^{u}_{t}$: the edge convolution starts learning an embedding based on the absolute position and velocity information of the node, as well as the relative position and speed of its neighbors, or their team membership and variation of appearance. This is quite relevant given our hypothesis of high inter-dependency of motions between agents in football. Then, layer by layer, the network progressively learn a more abstract and more discriminative node embedding, simply guided by the loss function during training.

\bigskip

We only use a few layers of edge convolutions: in practice 3 or 4, to avoid the "over-smoothing" effect of the nodes.

\subsection{Final prediction}

Finally, we concatenate $\Phi_{X3D} \in \mathbb{R}^{NxTxD}$ with the tensor of nodes features obtained after application of the last layer K of graph edge convolution $h^{(K)} \in \mathbb{R}^{NxTxD'}$, apply the temporal convolution network of the TAAD and produce a dense tensor of predictions using a using a Multi-layer Perceptron (MLP). We train our network in a supervised manner using a cross entropy loss function (CE-Loss).

\subsection{Action tube construction}

An event is defined by its list of characteristics: (start frame, end frame, class, score). Both TAAD and our method output dense predictions (class prediction per frame and per player), that can contain spurious detections on a few frames in the middle of a sequence of the same label. 

\bigskip

Therefore, in order to transform this sequence into an action-tube we filter it by performing a label smoothing optimization \cite{Saha_2016,Singh_2017,kalogeiton_action_2017}. Only then the sequence of events is extracted and matched with the ground truth labels to calculate our metrics. A predicted event is added to the list of events when its class is not background.

\subsection{Implementation details}

We used 50 frames as input with no sub-sampling (2 seconds of video clip). We used the X3D S pre-trained backbone as feature extractor and the PyTorch Geometric framework was used for the Edge Convolutions layers. We used the Adam optimizer \cite{kingma_adam_2017}, with a learning rate of 0.0005, a weight decay of 1e-05 on the non-bias parameters, a gradient accumulation over 20 iterations before weight update and a batch size of 6 on a RTX A6000 GPU. We trained the system end-to-end for a total of 13 epochs and divided the learning rate by 10 at epoch 10.

\section{\uppercase{Evaluation}}

\subsection{Metrics}

We have chosen to report the mean Average Precision as our main metric. Is is common in object detection and video action detection works \cite{Weinzaepfel_2015,kalogeiton_action_2017}. A detection is correct when its Intersection-Over-Union (IoU) with a ground truth action tube is larger than a given threshold (e.g. 0.5 or 0.2) and the predicted label matches the ground truth one. Then, Average Precision is computed for each class, using the 11 points method as proposed by \cite{PascalVOC2010}, and the mean is computed across classes. The particularity of our setup is that the tracking of the players is already done: therefore, the IoU is purely temporal.

\bigskip

In more details, the matching between predicted and ground truth events is performed as follows: the temporal IoU is calculated for all event pairs to identify the best match, regardless of class. If the highest IoU exceeds the threshold, the events are matched and removed from their lists. Matches with identical classes are True Positives. Unmatched true events (he has not been detected) are paired with "dummy" predicted events of the background class, and unmatched predicted events are paired with "dummy" true events of the background class.

\subsection{Comparisons}

We implemented TAAD and TAAD + GNN as described in the Methdology section. 

\bigskip

\textbf{Firstly}, we observe that metrics at 0.2 IoU threshold obtained when running our implementation of TAAD on the Footovision dataset are close to the metrics reported by the authors on their implementation of TAAD on the MultiSports dataset. Though these results cannot be compared directly (there can be slight domain gaps, the backbones are not identical, etc.), this shows that this method translates well to this new dataset, even with a lighter backbone:

\begin{itemize}
    \item \textit{60.6\% mAP@0.2 and 37\% mAP@0.5 on MultiSports}
    \item \textit{59.58\% mAP@0.2 and 46.4\% mAP@0.5 on the Footovision dataset}
\end{itemize}

The gap observed on the mAP at a 0.5 threshold of IoU could be explained by the fact that our dataset is less challenging than MultiSports: less classes, less class imbalance, more samples per class on the minority classes.

\bigskip

\begin{figure}[h!]\centering
\includegraphics[width=.99\linewidth]{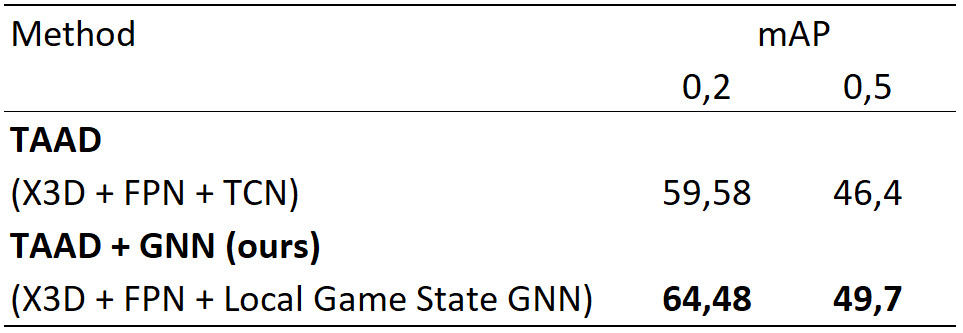}
\caption{Comparison of TAAD and our method TAAD + GNN: integrating structured game state information into the detection process leads to improvements in performances.}
\label{fig:comp_taad_taadgnn}
\end{figure}

\begin{figure}[h!]\centering
\includegraphics[width=.99\linewidth]{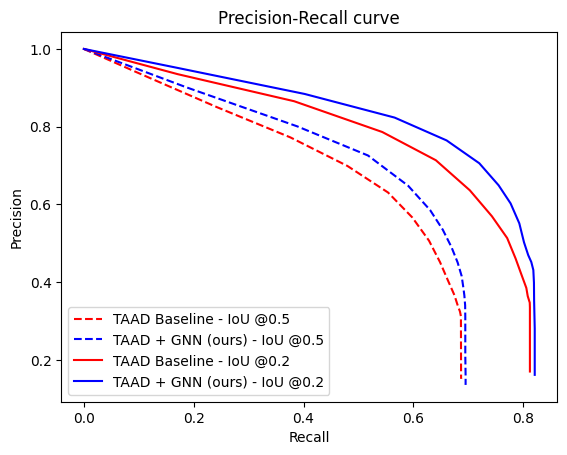}
\caption{Comparison of the Precision - Recall curves for the TAAD and TAAD + GNN, with various IoU thresholds (0.2 and 0.5)}
\label{fig:PR_REC_curves}
\end{figure}

\textbf{Secondly}, Figure \ref{fig:comp_taad_taadgnn} and Figure \ref{fig:PR_REC_curves} show the higher performance of TAAD + GNN compared to TAAD alone. This demonstrates that integrating structured game state information into the detection process leads to improvements in detections.

\subsection{Discussion}

\subsubsection{High recall and low precision regime}

In order to meet the expectation of exhaustiveness of events detection in the context of data analytics, it is envisaged to work in a regime of high recall - low precision, using predictions with low confidence scores. Though this setting unavoidably leads to the detection of numerous False Positives, their subsequent filtering by human intervention remains far more competitive compared to the alternative: seeking missing information by "scrubbing" a video.

\bigskip

 We assess the interest of our method in this context of high recall - low precision by studying the variations in the number of True Positives and False Positives when switching from TAAD to TAAD + GNN. For that purpose, we fixed the IoU threshold at 0.2 and the confidence score threshold at 0.5 (for an overall recall reaching $\sim$ 80\%).
 
 \bigskip
 
\begin{figure}[h]\centering
\includegraphics[width=.99\linewidth]{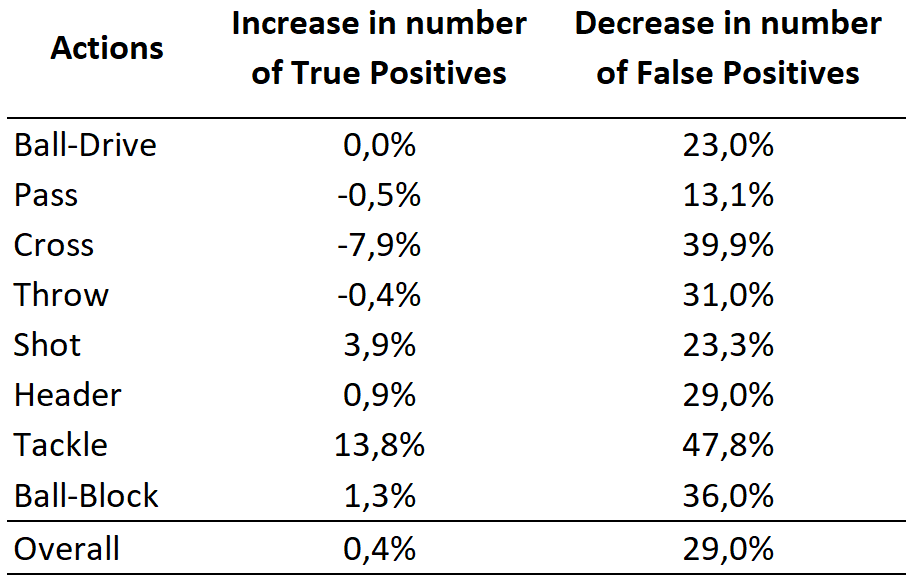}
\caption{When working in high recall - low precision regime, our method significantly reduce the number of False Positive detections}
\label{fig:high_recall_low_precision}
\end{figure}

 The results are reported in Figure \ref{fig:high_recall_low_precision} and allow us to confirm the benefit of our method in this context: making more structured predictions by using game state information reduces the number of false positives by almost 30\%, while keeping the overall number of True Positives constant.

\subsubsection{Detailed variations of performances}

A more detailed analysis of the results shows that despite an overall improvement in the metrics, the situation is quite contrasted on a per class level. 

\bigskip

\begin{figure}[h]\centering
\includegraphics[width=0.99\linewidth]{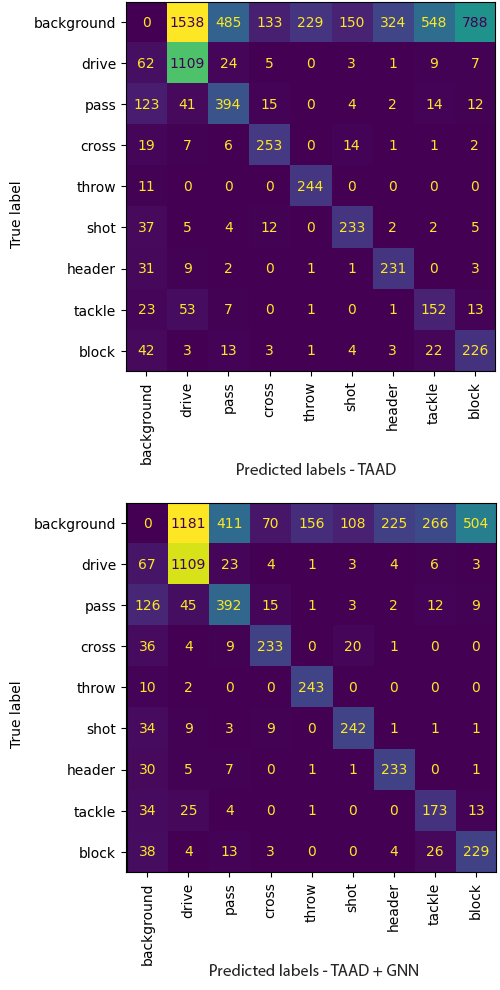}
\caption{Confusion matrices obtained when applying TAAD and TAAD + GNN, with IoU threshold at 0.2 and confidence score threshold at 0.5}
\label{fig:confmat}
\end{figure}

Indeed, analyzing the class level results reported in Figure \ref{fig:high_recall_low_precision}, we observe an important erosion of detection for the class "Cross" (almost -8\%). It is partly explained by the higher confusion between "Cross" actions and "Shot" actions when switching from TAAD to TAAD + GNN, while, surprisingly, less "Shots" are confused with "Crosses" (see the confusion matrices in Figure \ref{fig:confmat}). These two classes can be easily confused: the players performing both actions are in the same zones of the pitch and both shoot the ball toward the goalmouth, except that in the first case the player intend to pass the ball to a player that is closer to the goalmouth, while in the second case he intends to score directly. It is also sometimes difficult to know what was the intention of the player when he misses one of those actions.

\bigskip

"Cross" is also the class that experience the second most important decrease in its number of False Positives. Therefore, a shift in the equilibrium recall-precision for the current confidence score threshold (0.5), between the two methods, could be an another factor of explanation.

\bigskip

Finally, the "Tackle class" experiences the most important improvement in both decrease of False Positives (almost 48\%) and increase in True Positives (almost 14\%). The game state information seems, on one hand, to limit the class confusions between "Tackle" and "Ball-drive" (around -52\%) and, on the other hand, reduce the level of False Positives, as seen in the confusion matrices in Figure \ref{fig:confmat}.

\subsubsection{The benefits of structured predictions}

In order to better understand in which cases structured prediction is beneficial, and explain the observed improvement in our metrics, we compiled, per class, the clips that contained false positive events when using the TAAD, but that didn't fool the TAAD + GNN approach.

\bigskip

One key benefit is the reduction of simultaneous false positives. TAAD occasionally detects actions that overlap with ongoing real actions, which is unrealistic given there is only one ball and one active player controlling it. For instance, TAAD might falsely detect a player on the sideline as performing a throw-in while another player is actively driving the ball, due to reliance solely on player appearance and gestures. By incorporating surrounding player information and their relative configurations, such false positives are eliminated, particularly in the Throw class. While filtering detections by score could be an alternative, 3D CNNs often exhibit overconfidence, with false positives sometimes scoring higher than the true action.

\bigskip

Another benefit is the reduction of false positives in actions like Ball-blocks or Header. False positives often occur when a player attempts the gesture but misses the ball, where motion blur and camera perspective falsely suggest contact. These actions usually happen in crowded areas, and successful execution typically causes a noticeable ball trajectory change, prompting surrounding players to accelerate. When the ball is missed, the players’ movement remains less disrupted. Our method appears to compensate for 3D CNNs' difficulty in linking ball trajectory to player contact by learning how groups of players react to sudden ball trajectory changes.

\subsubsection{Team sports and occlusion}

Many false positives happen when players are really close to each others, leading either to ambiguous situations, when two players are equidistant from the ball in screen space, or even to impossible detections when players are not visible. Sometimes, two players can have their bounding boxes with very high overlap, which means that their features, extracted using a ROI Align layer, are almost identical. This is for example the case in many Header events when two players are performing aerial duels. For those hard cases, further research will be necessary, and it is highly probable that each class of event will require its own approach.

\section{\uppercase{Conclusions}}
\label{sec:conclusion}

We propose a method that combines visual features and structured game state information, using Graph Neural Networks, to enhance more traditional spatio-temporal action detection methods in soccer. We demonstrate the benefit of our method in the context of high recall - low precision regime, required to generate reliable statistics more efficiently. Nevertheless, our method sees little of the game: only 2 seconds. In future works we envisage to push our hypothesis further and overcome this temporal horizon by (i) using our method to make “prior” predictions of actions for each player, (ii) use sequence to sequence methods to learn the logic and dynamic of the game on long sequences of events and leverage that knowledge to either generate "posterior" predictions per player, or denoise or score our sequences of detections, conditionally to the game state, and (iii) introduce additional game state features, traditionally found in the literature about game state based methods to predict future events.

\bibliographystyle{apalike}
{\small

\end{document}